\documentstyle[aclap]{article}
\title{\vspace{-0.5in}Translation Methodology in the Spoken Language
Translator: \\ An Evaluation}

\author{\bf David Carter\\ 
        \bf Ralph Becket\\
        \bf Manny Rayner\\
        \ \\
        SRI International\\
        Suite 23, Millers Yard\\
        Cambridge CB2 1RQ\\
        United Kingdom\\
        \ \\
        {\tt dmc@cam.sri.com}\\
        {\tt rwab1@cam.sri.com}\\
        {\tt manny@cam.sri.com}\\
        \And
        \bf Robert Eklund\\
        \bf Catriona MacDermid\\
        \bf Mats Wir\'en\\
        \ \\
        Telia Research AB\\
        Spoken Language Processing\\
        S-136\,80 Haninge\\
        Sweden\\
        \ \\
       {\tt Robert.H.Eklund@telia.se}\\
       {\tt Catriona.I.Macdermid@telia.se}\\
       {\tt Mats.G.Wiren@telia.se}\\
       \And
        \bf Sabine Kirchmeier-Andersen\\
        \bf Christina Philp\\
        \ \\
        \ \\
        Handelsh\o jskolen i K\o benhavn\\
        Institut for Datalingvistik\\
        Dalgas Have 15\\
        DK-2000 Frederiksberg\\
        Denmark\\
        \ \\
        {\tt sabine.id@cbs.dk} \\
        {\tt cp.id@cbs.dk}}

\begin{document}
\maketitle
\bibliographystyle{fullname}

\vspace{-0.5in}
\begin{abstract}
In this paper we describe how the translation methodology adopted for
the Spoken Language Translator (SLT) addresses the characteristics of
the speech translation task in a context where it is essential to
achieve easy customization to new languages and new domains. We then
discuss the issues that arise in any attempt to evaluate a speech
translator, and present the results of such an evaluation carried out
on SLT for several language pairs.
\end{abstract}

\section{The nature of the speech translation task}

Speech translation is in many respects a particularly difficult
version of the translation task. High quality output is essential: the
speech produced must sound natural if it is to be easily
comprehensible. The quality of the translation itself must also be
high, in spite of the fact that, by the nature of the problem, no
post-editing is possible. Things are equally difficult on the input
side: pre-editing, too, is difficult or impossible, yet ill-formed
input and recognition errors are both likely to be quite common. Thus
robust analysis and translation are also required. Furthermore, any
attempted solutions to these problems must be capable of operating at
a speed close enough to real time that users are not faced with
unacceptable delays.

Together, these factors mean that speech translation is currently only
practical for limited domains, typically involving a vocabulary of a
few thousand words. Because of this, it is desirable that a speech
translator should be easily portable to new domains.  Portability to
new languages, involving the acquisition of both monolingual and
cross-linguistic information, should also be as straightforward as
possible. These ends can be achieved by using general-purpose
components for both speech and language processing and training them
on domain-specific speech and text corpora. The training should be
automated whenever possible, and where human intervention is required,
the process should be deskilled to the level where, ideally, it can be
carried out by people who are familiar with the domain but are not
experts in the systems themselves.

These points will be discussed in the context of the Spoken Language
Translator (SLT) \cite{SLT-HLT,SLT-report,RaynerCarter:97}, a
customizable speech translator built as a pipelined sequence of
general-purpose components. These components are: a version of the
Decipher (TM) speech recognizer \cite{Murveit:93} for the source
language; a copy of the Core Language Engine (CLE) \cite{CLE} for the
source language; another copy of the CLE for the target language; and
a target language text-to-speech synthesizer.  

The current SLT system carries out multi-lingual speech translation in
near real time in the ATIS domain \cite{ATIS} for several language
pairs. Good demonstration versions exist for the four pairs English
$\rightarrow$ Swedish, English $\rightarrow$ French, Swedish
$\rightarrow$ English and Swedish $\rightarrow$ Danish. Preliminary
versions exist for five more pairs: Swedish $\rightarrow$ French,
French $\rightarrow$ English, English $\rightarrow$ Danish, French
$\rightarrow$ Spanish and English $\rightarrow$ Spanish.

We describe the methodology used to build the SLT system itself,
particularly in the areas of customization (Section
\ref{customization}), robustness (Section \ref{robustness}), and
multilinguality (Section \ref{multiling}). For further details on the
topics of customization and multilinguality, see
\cite{RaynerBretanEtAl:96,RaynerCarterEtAl:97}; and on robustness, see
\cite{RaynerCarter:97}. We then discuss the evaluation of speech
translation systems. This is an area that deserves more attention than
it has received to date; indeed, it is not obvious how best to perform
such an evaluation so as to measure meaningfully the performance both
of the overall system and of each of its components. In Sections
\ref{Section:Eval-methodology} and \ref{Section:SLT-evaluation} of
this paper, we therefore consider the characteristics an evaluation
should have, and describe one we have carried out, discussing the
extent to which it meets the desired criteria.

\section{Customization to languages and domains}
\label{customization}

In the Core Language Engine, the language processing component of the
Spoken Language Translator system, we address the requirement of
portability by maintaining a clear separation between (1) the system
code; (2) linguistic rules, including lexicon entries, to generate
possible analyses and translations non-deterministically; and (3)
statistical information, to choose between these possibilities.  The
practical advantage of this architecture is that most of the work
involved in porting the system to a new domain is concerned with the
parts of the system that can be modified by non-experts: the central
activities are addition of new lexicon entries, and supervised
training to derive the statistical preference information.  Porting to
new languages is a more complex task, but still only involves
modifications to a relatively small subset of the whole system. In
more detail:

\vspace{0.1in} \noindent {\bf (1)} The {\it system code} is completely
general-purpose and does not need any changes for new domains or,
other than in exceptional cases,\footnote{E.g.\ in our initial
extension from English to languages with more complicated morphology,
which necessitated the development of a morphological processor based
on the two-level formalism (see \cite{Carter:95}).} for new
languages.

\vspace{0.1in} \noindent 
{\bf (2)} The more complex of the {\it linguistic rules} for a given
language are the grammar, the function word lexicon, and the macros
defining common content word behaviours (count noun, transitive verb,
etc). These are defined using explicit feature-value equations which
must be written by a skilled grammarian.  For a given language {\it
pair}, the more complex transfer rules, which tend to be for function
words and other commonly-occurring, idiosyncratic words, can also
involve arbitrarily large, recursive structures.  However, nearly all
of these monolingual and bilingual rules are domain-independent.

On the other side of the coin, the main domain-dependent aspects of a
linguistic description are lexicon entries defining content words in
terms of existing behaviours, and simple (atomic-to-atomic) transfer
rules. These do need to be created manually for each new domain, but
they are simple enough to be defined by non-experts with the help of
relatively simple graphical tools. See Figures 1 and 2 for some
examples of these two kinds of rule (the details of the formalism are
unimportant here, we intend simply to illustrate the differences in
complexity).


\begin{figure*}
Syntax rule for S $\rightarrow$ NP VP:
\begin{verbatim}
   syn(s_np_vp_Normal, core,
    [s:[@s_np_feats(MMM), @vp_feats(MM), 
        sententialsubj=SS,sai=Aux, hascomp=n,conjoined=n], 
      np:[@s_np_feats(MMM),vform=(fin\/to), relational=_,temporal=_,agr=Ag,
          sentential=SS, wh=_, whmoved=_,pron=_,nform=Sfm],
      vp:[@vp_feats(MM),vform=(\(en)),agr=Ag,sai=Aux, modifiable=_,  
          mainv=_,headfinal=_,subjform=Sfm]]).
\end{verbatim}
Macro definition for syntax of transitive verb:
\begin{verbatim}
   macro(v_subj_obj,
         [v:[vform=base,mhdfl=A,passive=A,gaps=B,conjoined=n,
             subcat=[np:[relational=_,passive=A,wh=_,gap=_,gaps=B,
                         temporal=_,pron=_,case=nonsubj]]]]).
\end{verbatim}
Transfer rule relating English adjective ``early'' and French PP ``de
bonne heure'':
\begin{verbatim}
   trule([eng,fre],semi_lex(early-de_bonne_heure),
         [early_NotLate,tr(arg)]
         ==
         @form(prep('de bonne heure_Early'),_,
               P^[P,tr(arg),
                    @term(ref(pro,de_bonne_heure,sing,_),
                    V,W^[time,W])+_])).
\end{verbatim}
\label{rulefig1}
\caption{Complex, domain-independent linguistic rules}
\end{figure*}

When moving to a new language, more expert intervention is typically
required than for a new domain, because many of the complex rules do
need some modifications. However, we have found that the amount of
work involved in developing new grammars for Swedish, French, Spanish
and most recently Danish has always been at least an order of
magnitude less than the effort required for the original grammar
\cite{GambackRayner:92,RaynerCarterBouillon:96,RaynerCarterEtAl:97}.

\vspace{0.1in} \noindent
{\bf (3)} The {\it statistical information} used in analysis
is entirely derived from the results of supervised training on corpora
carried out using the TreeBanker \cite{Carter:97}, a graphical tool
that presents a non-expert user with a display of the salient
differences between alternative analyses in order that the correct one
may be identified.  Once a user has become accustomed to the system,
around two hundred sentences per hour may be processed in this way.
This, together with the use of representative subcorpora
\cite{RaynerBouillonCarter:95} to allow structurally equivalent sentences to be
represented by a single example, means that a corpus of many thousands
of sentences can be judged in just a few person weeks. The principal
information extracted automatically from a judged corpus is:
\begin{itemize}
\item Constituent pruning rules, which allow the detection and
removal, at intermediate stages of parsing, of syntactic constituents
occurring in contexts where they are unlikely to contribute to the
correct parse. Removing these constituents significantly constrains
the search space and speeds up parsing \cite{RaynerCarter:97}.
\item An automatic tuning of the grammar to the domain using the
technique of Explanation-Based Learning
\cite{HarmelenBundyEBL,FGCS-88,EBL-IJCAI-91,ACL96}. This rewrites it
to a form where only commonly-occurring rule {\it combinations} are
represented, thus reducing the search space still further and giving
an additional significant speedup.
\item Preference information attached to certain characteristics of
full analyses of sentences -- the most important being semantic
triples of head, relationship and modifier -- which allow a selection
to be made between competing full analyses. See
\cite{AlshawiCarter:94} and \cite{Carter:97} for details.
\end{itemize}
A similar mechanism has been developed to allow users to specify
appropriate translations, giving rise to preferences on outcomes of
the transfer process. Work on this continues.

\begin{figure*}
Lexicon entry, using transitive verb macro, for ``serve'' as in
``Does Continental serve Atlanta?'':
\begin{verbatim}
   lr(serve,v_subj_obj,serve_FlyTo).
\end{verbatim}
Transfer rule relating that sense of ``serve'' to one sense of
French ``desservir'':
\begin{verbatim}
   trule([eng,fre],lex(simple),serve_FlyTo==desservir_ServeCity).
\end{verbatim}
\label{rulefig2}
\caption{Simple, domain-dependent linguistic rules}
\end{figure*}

\section{Robustness}
\label{robustness}

Robustness in the face of ill-formed input and recognition errors is
tackled by means of a ``multi-engine'' strategy
\cite{FrederkingNirenburg:94,RaynerCarter:97}, combining two different
translation methods. The main translation method uses transfer at the
level of QLF \cite{AlshawiEtAl:91,RaynerBouillon:95}; this is
supplemented by a simpler, glossary-based translation
method. Processing is carried out bottom-up. Roughly speaking, the QLF
transfer method is used to translate as much as possible of the input
utterance, any remaining gaps being filled by application of the
glossary-based method.

In more detail, source-language parsing goes through successive stages
of lexical (morphological) analysis, low-level phrasal parsing to
identify constituents such as simple noun phrases, and finally full
sentential parsing using a version of the original grammar tuned to
the domain using explanation-based learning (see Section
\ref{customization} above).  Parsing is carried out in a bottom-up
mode. After each parsing stage, a corresponding translation operation
takes place on the resulting constituent lattice. Translation is
performed by using the glossary-based method at the early stages of
processing, before parsing is initiated, and by using the QLF-transfer
method during and after parsing. Each successful transfer attempt
results in a target language string being added to a target-side
lattice.  Metrics are then applied to choose a path through this
lattice. The criteria used to select the path involve preferences for
sequences that have been encountered in a target-language corpus; for
the use of more sophisticated transfer methods over less
sophisticated; and for larger over smaller chunks.

The bottom-up approach contributes to robustness in the obvious way:
if a single analysis cannot be found for the whole utterance, then
translations can be produced for partial analyses that have already
been found. It also contributes to system response in that the
earlier, more local, shallower methods of analysis and transfer
usually operate very quickly to produce an attempt at translation. The
target-language user may interrupt processing before the more global
methods have finished if the translation (assuming it can be viewed on
a screen) is adequate, or the system itself may abandon a sentence,
and present its current best translation, if a specified time has
elapsed.

Figure \ref{robustex} exemplifies the operation of the multi-engine
strategy as well as of the preferences applied to analysis and
transfer.\footnote{The example chosen was the most interesting of the
dozen or so in our most recent demonstration session, and the
intermediate results have been reproduced from the system log file
without any changes other than reformatting.}  The N-best list
delivered by the speech recognizer contains the sentence actually
uttered, ``Could you show me an early flight please?'', but only in
fourth position.
\begin{itemize}
\item Before any linguistic processing is carried out, the word
sequence at the top of the N-best list is the most preferred one, as
only recognition preferences (shown by position in the list) are
available. This sequence is translated word-for-word using the
glossary method, giving result (a) in the figure.
\item After lexical analysis, which effectively includes
part-of-speech tagging, it is determined that the word ``a'' is
unlikely to precede ``are'', and so ``a'' is dropped from the
translated sequence (b) -- thus translating recognizer hypothesis 2,
using the glossary-based method.
\item Phrasal parsing identifies ``an early flight'' as a likely noun
phrase, so that this is for the first time selected for translation,
in (c). Note that the system has now settled on the correct English
word sequence.  QLF-based transfer is used for the first time, and the
transfer rule in Figure 1 
is used to translate ``early'' as ``de bonne heure'' which, because it
is a PP, is placed after ``vol'' (flight) by the French grammar.
\item Finally, as shown in (d), an analysis and a QLF-based
translation are found for the whole sentence, allowing the inadequate
word-for-word translation of ``could you show me'' as ``*pourriez vous
montrez moi'' to be improved to a more grammatical ``pourriez-vous
m'indiquer''.
\end{itemize}
We thus see the results of translation becoming steadily more accurate
and comprehensible as processing proceeds.

\begin{figure*}
N-best list (N=5) delivered by speech recognizer:
\begin{verbatim}
   1  could you show me a are the flight please
   2  could you show me are the flight please
   3  could you show me in order a flight please
   4  could you show me an early flight please
   5  could you show meals are the flight please
\end{verbatim}

(a) Selected input sequence and translation after surface phase:

\vspace{2mm}
\begin{tabular}{|c|c|c|c|c|c|c|c|c|}\hline
could & you & show & me & a & are & the & flight & please \\ \hline
pourriez & vous & montrez & moi & un & sont & les & vol & s'il vous
pla\^{\i}t \\ \hline
\end{tabular}
\vspace{4mm}

(b) Selected input sequence and translation after lexical phase:

\vspace{2mm}
\begin{tabular}{|c|c|c|c|c|c|c|c|}\hline
could & you & show & me & are & the & flight & please \\ \hline
pourriez & vous & montrez & moi & sont & les & vol & s'il vous pla\^{\i}t
\\ \hline
\end{tabular}
\vspace{4mm}

(c) Selected input sequence and translation after phrasal phase:

\vspace{2mm}
\begin{tabular}{|c|c|c|c|c|c|}\hline
could & you & show & me & an early flight & please  \\ \hline
pourriez & vous & montrez & moi & un vol de bonne heure & s'il vous
pla\^{\i}t \\ \hline 
\end{tabular}
\vspace{4mm}

(d) Selected input sequence and translation after full parsing phase:

\vspace{2mm}
\begin{tabular}{|c|}\hline
could you show me an early flight please \\ \hline
pourriez-vous m'indiquer un vol de bonne heure s'il vous pla\^{\i}t
\\ \hline
\end{tabular}
\label{robustex}
\caption{N-best list and translation results for ``Could you show me
an early flight please?''}
\end{figure*}

\section{Multilinguality,  interlinguas and the ``N-squared problem''}
\label{multiling}

While using an interlingual representation would seem to be the
obvious way to avoid the ``N-squared problem'' (translating between
$N$ languages involves order $N^2$ transfer pairs), we are sceptical
about interlinguas for the following reasons. 

Firstly, doing good translation is a mixture of two tasks: semantics
(getting the meaning right) and collocation (getting the appearance of
the translation right). Defining an interlingua, even if it is
possible to do so for an increasing number $N$ of languages, really
only addresses the first task. Interlingual representations also tend
to be less portable to new domains, since they if they are to be
truly interlingual they normally need to be based on domain concepts,
which have to be redefined for each new domain -- a task that involves
considerable human intervention, much of it at an expert level.
In contrast, a transfer-based representation can be shallower (at the
level of linguistic predicates) while still abstracting far enough
away from surface form to make most of the transfer rules simple
atomic substitutions.

Secondly, systems based on formal representations are brittle: a fully
interlingual system first needs to translate its input into a formal
representation, and then realise the representation as a
target-language string. An interlingual system is thus inherently
more brittle than a transfer system, which can produce an output
without ever identifying a ``deep'' formal representation of the
input. For these reasons, we prefer to stay with a fundamentally
transfer-based methodology; none the less, we include some
aspects of the interlingual approach, by regularizing the
intermediate QLF representation to make it as language-independent
as possible consonant with the requirement that it also be independent
of domain. Regularizing the representation has the positive effect of 
making the transfer rules simpler (in the limiting case, a fully
interlingual system, they become trivial). 

We tackle the N-squared problem by means of {\it transfer composition}
\cite{RaynerCarterBouillon:96,RaynerCarterEtAl:97}. If we already have
transfer rules for mapping from language $A$ to language $B$ and from
language $B$ to language $C$, we can compose them to generate a set to
translate directly from $A$ to $C$. The first stage of this
composition can be done automatically, and then the results can be
manually adjusted by adding new rules and by introducing declarations
to disallow the creation of implausible rules: these typically arise
because the contexts in which $\alpha \in A$ can correctly be
translated to $\beta \in B$ are disjoint from those in which $\beta$
can be translated into $\gamma \in C$.  As with the other
customization tasks described here, the amount of human intervention
required to adjust a composed set of transfer rules is vastly less,
and less specialized, than what would be required to write them from
scratch.

In the current version of SLT, transfer rules were written directly
for neighbouring languages in the sequence Spanish -- French --
English -- Swedish -- Danish (most of these neighbours being
relatively closely related), with other pairs being derived by
transfer composition. Further details can be found in
\cite{RaynerCarterEtAl:97}.

\section{Evaluation of speech translation systems: methodological issues}
\label{Section:Eval-methodology}

There is still no real consensus on how to evaluate speech translation
systems.  The most common approach is some version of the following.
The system is run on a set of previously unseen speech data; the
results are stored in text form; someone judges them as acceptable or
unacceptable translations; and finally the system's performance is
quoted as the proportion that are acceptable. This is clearly much
better than nothing, but still contains some serious methodological
problems. In particular:

\begin{enumerate}

\item There is poor agreement on what constitutes an ``acceptable
translation''. Some judges regard a translation as unacceptable if
a single word-choice is suboptimal. At the other end of the scale,
there are judges who will accept any translation which conveys the
approximate meaning of the sentence, irrespective of how many
grammatical or stylistic mistakes it contains. Without specifying more
closely what is meant by ``acceptable'', it is difficult to compare
evaluations.

\item Speech translation is normally an interactive process, and it is
natural that it should be less than completely automatic. At a
minimum, it is clearly reasonable in many contexts to feed back to the
source-language user the words the recognizer believed it heard, and
permit them to abort translation if recognition was unacceptably
bad. Evaluation should take account of this possibility.

\item Evaluating a speech-to-speech system as though it were a
speech-to-text system introduces a certain measure of
distortion. Speech and text are in some ways very different media: a
poorly translated sentence in written form can normally be re-examined
several times if necessary, but a spoken utterance may only be heard
once. In this respect, speech output places heavier demands on
translation quality. On the other hand, it can also be the case that
constructions which would be regarded as unacceptably sloppy in
written text pass unnoticed in speech.

\end{enumerate}

We are in the process of redesigning our translation evaluation
methodology to take account of all of the above points. Currently,
most of our empirical work still treats the system as though it
produced text output; we describe this mode of evaluation in
Section~\ref{Section:Speech-to-text-eval}. A novel method which
evaluates the system's actual spoken output is currently undergoing
initial testing, and is described in
Section~\ref{Section:Speech-to-speech-eval}. Section~\ref{Section:SLT-evaluation}
presents results of experiments using both evaluation methods.

\subsection{Evaluation of speech to text translation}
\label{Section:Speech-to-text-eval}

In speech-to-text mode, evaluation of the system's performance on a
given utterance proceeds as follows. The judge is first shown a text
version of the correct source utterance (what the user actually said),
followed by the selected recognition hypothesis (what the system
thought the user said). The judge is then asked to decide whether the
recognition hypothesis is acceptable. Judges are told to assume that
they have the option of aborting translation if recognition is of
insufficient quality; judging a recognition hypothesis as unacceptable
corresponds to pushing the `abort' button.

When the judge has determined the acceptability of the recognition
hypothesis, the text version of the translation is presented. (Note
that it is not presented earlier, as this might bias the decision
about recognition acceptability.) The judge is now asked to classify
the quality of the translation along a seven-point scale; the points
on the scale have been chosen to reflect the distinctions judges most
frequently have been observed to make in practice. When selecting the
appropriate category, judges are instructed only to take into account
the actual spoken source utterance and the translation produced, and
ignore the recognition hypothesis. The possible judgement categories
are the following; the headings are those used in Tables
\ref{Table:speech-to-text-auto} and
\ref{Table:speech-to-text-with-abort} below.

\begin{description}

\item[Fully acceptable.] Fully acceptable translation.

\item[Unnatural style.] Fully acceptable, except that style is not completely
natural. This is most commonly due to over-literal translation.

\item[Minor syntactic errors.] One or two minor syntactic or word-choice errors,
otherwise acceptable. Typical examples are bad choices of determiners
or prepositions.

\item[Major syntactic errors.] At least one major or several minor
syntactic or word-choice errors, but the sense of the utterance is
preserved. The most common example is an error in word-order produced
when the system is forced to back up to the robust translation method.

\item[Partial translation.] At least half of the utterance has been acceptably
translated, and the rest is nonsense. A typical example is when
most of the utterance has been correctly recognized and translated, 
but there is a short `false start' at the beginning which has resulted
in a word or two of junk at the start of the translation.

\item[Nonsense.] The translation makes no sense. The most common reason
is gross misrecognition, but translation problems can sometimes be
the cause as well.

\item[Bad translation.] The translation makes some sense, but fails to
convey the sense of the source utterance. The most common reason is
again a serious recognition error.

\end{description}

Results are presented by simply counting the number of translations in
a run which fall into each category. By taking account of the
``unacceptable hypothesis'' judgements, it is possible to evaluate the
performance of the system either in a fully automatic mode, or in a
mode where the source-language user has the option of aborting
misrecognized utterances.

\subsection{Evaluation of speech to speech translation}

\label{Section:Speech-to-speech-eval}

Our intuitive impression, based on many evaluation runs in several
different language-pairs, is that the ``fine-grained'' style of
speech-to-text evaluation described in the preceding section gives a
much more informative picture of the system's performance than the
simple acceptable/unacceptable dichotomy. However, it raises an
obvious question: how important, in objective terms, are the
distinctions drawn by the fine-grained scale? The preliminary work we
now go on to describe attempts to provide an empirically justifiable
answer, in terms of the relationship between translation quality and
comprehensibility of output speech.  Our goal, in other words, is to
measure objectively the ability of subjects to understand the content
of speech output. This must be the key criterion for evaluating a
candidate translation: if apparent deficiencies in syntax or
word-choice fail to affect subject's ability to understand content,
then it is hard to say that they represent real loss of quality.

The programme sketched above is difficult or, arguably, impossible to
implement in a general setting. In a limited domain, however, it
appears quite feasible to construct a domain-specific form-based
questionnaire designed to test a subject's understanding of a given
utterance. In the SLT system's current domain of air travel planning
(ATIS), a simple form containing about 20 questions extracts enough
content from most utterances that it can be used as a reliable measure
of a subject's understanding. The assumption is that a normal domain
utterance can be regarded as a database query involving a limited
number of possible categories: in the ATIS domain, these are concepts
like flight origin and destination, departure and arrival times,
choice of airline, and so on. A detailed description of the evaluation
method follows.

The judging interface is structured as a hypertext document that can
be accessed through a web-browser. Each utterance is represented by
one web page. On entering the page for a given utterance, the judge
first clicks a button that plays an audio file, and then fills in an
HTML form describing what they heard. Judges are allowed to start by
writing down as much as they can of the utterance, so as to keep it
clear in their memory as they fill in the form.

The form is divided into four major sections. The first deals with the
linguistic form of the enquiry, for example, whether it is a command
(imperative), a yes/no-question or a wh-question. In the second
section the judge is asked to write down the principal ``object'' of
the utterance. For example, in the utterance ``Show flights from
Boston to Atlanta'', the principal object would be ``flights''.  The
third section lists some 15 constraints on the object explicitly
mentioned in the enquiry, like ``\ldots one-way from New York to
Boston on Sunday''. Initial testing proved that these three sections
covered the form and content of most enquiries within the domain, but
to account for unforeseen material the judge is also presented with a
``miscellaneous'' category. Depending on the character of the options,
form entries are either multiple-choice or free-text. All form entries
may be negated (``No stopovers'') and disjunctive enquiries are
indicated by dint of indexing (``Delta on Thursday or American on
Friday''). When the page is exited, the contents of the completed form
are stored for further use.

Each translated utterance is judged in three versions, by different
judges. The first two versions are the source and target speech files;
the third time, the form is filled in from the {\it text} version of
the source utterance. (The judging tool allows a mode in which the
text version is displayed instead of an audio file being played.) The
intention is that the source text version of the utterance should act
as a baseline with which the source and target speech versions can
respectively be compared. Comparison is carried out by a fourth judge.
Here, the contents of the form entries for two versions of the
utterance are compared.  The judge has to decide whether the contents
of each field in the form are compatible between the two versions.

When the forms for two versions of an utterance have been filled in
and compared, the results can be examined for comprehensibility in
terms of the standard notions of precision and recall.  We say that
the recall of version 2 of the utterance with respect to version 1 is
the proportion of the fields filled in version 1 that are filled in
compatibly in version 2\@. Conversely, the precision is the proportion
of the fields filled in in version 2 that are filled in compatibly in
version 1\@.

The recall and precision scores together define a two-element vector
which we will call the {\it comprehensibility\/} of version 2 with
respect to version 1\@. We can now define $C_{source}$ to be the
comprehensibility of the source speech with respect to the source
text, and $C_{target}$ to be the comprehensibility of the target
speech with respect to the source text. Finally, we define the quality
of the translation to be $1 - (C_{source} - C_{target})$, where
$C_{source} - C_{target}$ in a natural way can be interpreted as the
extent to which comprehensibility has degraded as a result of the
translation process.  At the end of the following section, we describe
an experiment in which we use this measure to evaluate the quality of
translation in the English $\rightarrow$ French version of SLT.

\section{An evaluation of the Spoken Language Translator}

\label{Section:SLT-evaluation}

We begin by presenting the results of tests run in speech-to-text mode
on versions of the SLT system developed for six different
language-pairs: English $\rightarrow$ Swedish, English $\rightarrow$
French, Swedish $\rightarrow$ English, Swedish $\rightarrow$ French,
Swedish $\rightarrow$ Danish, and English $\rightarrow$ Danish. Before
going any further, it must be stressed that the various versions
of the system differ in important ways; some language-pairs are
intrinsically much easier than others, and some versions of the system
have received far more effort than others.

In terms of difficulty, Swedish $\rightarrow$ Danish is clearly the
easiest language-pair, and Swedish $\rightarrow$ French is clearly the
hardest. English $\rightarrow$ French is easier than Swedish
$\rightarrow$ French, but substantially more difficult than any of the
others. English $\rightarrow$ Swedish, Swedish $\rightarrow$ English
and English $\rightarrow$ Danish are all of comparable difficulty.  We
present approximate figures for the amounts of effort devoted to each
language pair in conjunction with the other results.

We evaluated performance on each language-pair in the manner described
in Section~\ref{Section:Speech-to-text-eval} above, taking as input
two sets of 200 recorded speech utterances each (one for English and
one for Swedish) which had not previously been used for system
development. Judging was done by subjects who had not participated in
system development, were native speakers of the target language, and
were fluent in the source language. Results are presented both for a
fully automatic version of the system
(Table~\ref{Table:speech-to-text-auto}), and for a version with a
simulated `abort' button
(Table~\ref{Table:speech-to-text-with-abort}).

Finally, we turn to a preliminary experiment which used the
speech-to-speech evaluation methodology from
Section~\ref{Section:Speech-to-speech-eval} above. A set of 200
previously unseen English utterances were translated by the system
into French speech, using the same kind of subjects as in the previous
experiments. Source-language and target-language speech was
synthesized using commercially available, state-of-the-art
synthesizers (TrueTalk from Entropics and CNETVOX from ELAN
Informatique, respectively). The subjects were only allowed to hear
each utterance once. The results were evaluated in the manner
described, to produce figures for comprehensibility of source and
target speech respectively. The figures are presented in
Table~\ref{speech-to-speech-eval}; we expect to be able to present a
more detailed discussion of their significance by the time of the
workshop.

In summary, we have improved the standard evaluation method for speech
translation by developing a feasible alternative with a more
fine-grained taxonomy of acceptability. In order to make the task of
evaluation more realistic, we have also created a method in which
instead of textual translations it is the spoken form that is
judged. This method is currently in embryonic form, but the pilot
experiment described here leads us to think that the method shows
promise for further development.

An interesting future task would be to investigate the significance of
various kinds of written-language translation errors in terms of
reducing comprehensibility of the spoken output. This would amount to
systematically comparing $C_{target}$ with results obtained in
speech-to-text evaluations, divided up according to error categories
such as those in our taxonomy.

\section*{Acknowledgements}

The Danish-related work reported here was funded by SRI International
and Handelsh\o jskolen i K{\o}\-ben\-havn. Other work was funded by
Telia Research AB under the SLT-2 project.  We would like to thank
Beata Forsmark, Nathalie Kirchmeyer, Carin Lindberg, Thierry Reynier
and Jennifer Spenader for carrying out judging tasks.

\begin{table*}
\caption{Translation results for six language pairs on 200 unseen
utterances, all utterances in test set counted. Note that in both
tables on this page, the ``effort'' figures refer specifically to
translation work for the language pair in question, and exclude work
on grammar and lexicon development for the individual languages.}
\begin{center}
\begin{tabular}{|l|c|c|c|c|c|c|} \hline \hline
Source language        & English & English & Swedish & Swedish& Swedish& English \\ \hline
Target language        & Swedish & French  & English & French & Danish & Danish  \\ \hline \hline
Effort (person-months) &  8--10  &  3--5   & 3--5 & 1--2  & 0.5 &$<$0.5\\ \hline \hline
Fully acceptable       &  46.0\% & 52.0\%  & 45.0\%  & 19.0\% & 36.5\% &  27.0\% \\ \hline
Unnatural style        &  14.0\% & 10.5\%  & 4.5\%   & 15.0\% & 0.0\% &  0.0\%  \\ \hline
Minor syntactic errors &  12.0\% & 3.5\%   & 12.0\%  & 13.0\% & 37.5\% &  28.0\% \\ \hline \hline
{\bf Clearly useful}   &{\bf 72.0\%}&{\bf 66.0\%}&{\bf 61.5\%}&{\bf 47.0\%}&{\bf 74.0}\% &{\bf 55.0\%}\\ \hline \hline

Major syntactic errors &  7.0\%  & 2.5\%   & 7.5\%   & 13.0\% & 0.0\% &  0.0\%  \\ \hline
Partial translation    &  6.5\%  & 11.5\%  & 14.5\%  & 17.5\% & 1.5\%  &  1.5\%  \\ \hline \hline
{\bf Borderline}             &  {\bf 13.5\%} & {\bf 14.0\%}  & {\bf 22.0\%}  & {\bf 30.5\%} & {\bf 1.5\%} &  {\bf 1.5\%}  \\ \hline \hline

Nonsense               &  7.5\%  & 13.0\%  & 10.5\%  & 18.0\% & 13.0\% &  30.5\% \\ \hline
Bad translation        &  5.0\%  & 5.5\%   & 4.5\%   & 3.5\%  & 9.0\% &  10.5\% \\ \hline 
No translation         &  2.0\%  & 1.5\%   & 1.5\%   & 1.0\%  & 2.5\% &  2.5\%  \\ \hline \hline
{\bf Clearly useless}  &  {\bf 14.5\%} & {\bf 20.0\%}  & {\bf 16.5\%}  & {\bf 22.5\%} & {\bf 24.5\%} &  {\bf 43.5\%} \\ \hline \hline
\end{tabular}
\end{center}
\label{Table:speech-to-text-auto}
\end{table*}

\begin{table*}
\caption{Translation results for six language pairs on 200 unseen utterances,
ignoring utterances judged as recognition failures.}
\begin{center}
\begin{tabular}{|l|c|c|c|c|c|c|} \hline \hline
Source language        & English & English & Swedish & Swedish& Swedish& English \\ \hline
Target language        & Swedish & French  & English & French & Danish & Danish  \\ \hline \hline
Effort (person-months) &  8--10  &  3--5   & 3--5 & 1--2  & 0.5 &$<$0.5\\ \hline \hline
Fully acceptable       &  55.8\% &  65.8\% & 60.7\%  & 23.1\% & 49.0\%  & 35.9\%  \\ \hline
Unnatural style        &  15.8\% &  12.9\% & 6.4\%   & 19.2\% & 0.0\%   & 0.0\%   \\ \hline
Minor syntactic errors &  12.1\% &  3.2\%  & 11.4\%  & 15.4\% & 38.1\%  & 35.9\%  \\ \hline \hline
{\bf Clearly useful}         &  {\bf 83.7\%} &  {\bf 81.9\%} & {\bf 78.5\%}  & {\bf 57.7\%} & {\bf 87.1\%} & {\bf 71.8\%}  \\ \hline \hline

Major syntactic errors &  7.9\%  &  2.6\%  & 10.0\%  & 12.8\% & 0.0\%   & 0.0\%   \\ \hline
Partial translation    &  2.4\%  &  5.8\%  & 5.0\%   & 14.1\% & 0.7\%   & 2.1\%   \\ \hline \hline
{\bf Borderline}             &  {\bf 10.3\%} &  {\bf 8.4\%}  & {\bf 15.0\%}  & {\bf 26.9\%} & {\bf 0.7\%} & {\bf 2.1\%}   \\ \hline \hline

Nonsense               &  3.0\%  &  4.5\%  & 2.9\%   & 11.5\% & 4.8\%   & 12.4\%  \\ \hline
Bad translation        &  1.2\%  &  3.2\%  & 2.1\%   & 2.6\%  & 5.4\%   & 11.7\%  \\ \hline
No translation         &  1.8\%  &  1.9\%  & 1.4\%   & 1.3\%  & 2.0\%   & 2.1\%   \\ \hline \hline
{\bf Clearly useless}  &  {\bf 6.0\%}  &  {\bf 9.6\%}  & {\bf 6.4\%}   & {\bf 15.4\%} & {\bf 12.2\%} & {\bf 26.2\%}  \\ \hline \hline
(Utterances ignored)   &  35     &  45     & 60      & 44     & 53     & 55      \\ \hline \hline
\end{tabular}
\end{center}
\label{Table:speech-to-text-with-abort}
\end{table*}

\begin{table*}
\caption{Relative comprehensibility of source and target speech
for English $\rightarrow$ French test on 200 unseen utterances.}
\begin{center}
\begin{tabular}{|l|c|c|c|c|} \hline \hline
           &  Source  & Target & Difference & Quality \\ \hline \hline
Precision  &  97.6\%  & 86.0\%   & 11.6\%  &  88.4\% \\ \hline
Recall     &  97.5\%  & 84.0\%   & 13.5\%  &  86.5\% \\ \hline
\end{tabular}
\end{center}
\label{speech-to-speech-eval}
\end{table*}

\end{document}